\documentclass[10pt,twocolumn,letterpaper]{article}

\usepackage[pagenumbers]{cvpr}      
\usepackage{pifont}
\usepackage{multirow,multicol}
\definecolor{cvprblue}{rgb}{0.21,0.49,0.74}
\usepackage[pagebackref,breaklinks,colorlinks,allcolors=cvprblue]{hyperref}

\def\paperID{XAI4CV-10} 
\def\confName{CVPR}
\def\confYear{2026}

\title{From Attribution to Action: A Human-Centered Application of Activation Steering}

\author{Tobias Labarta\\
Fraunhofer Heinrich-Hertz-Institut\\
\and
Maximilian Dreyer\\
Fraunhofer Heinrich-Hertz-Institut\\
\and
\and 
Katharina Weitz\\
Fraunhofer Heinrich-Hertz-Institut\\
\and
Wojciech Samek\\
Fraunhofer Heinrich-Hertz-Institut\\
Technische Universität Berlin\\
BIFOLD – Berlin Institute for the Foundations of Learning and Data\\
\and
Sebastian Lapuschkin\\
Fraunhofer Heinrich-Hertz-Institut\\
}

\begin{document}
\maketitle

\begin{abstract}
Explainable AI (XAI) methods reveal which features influence model predictions, yet provide limited means for practitioners to act on these explanations. Activation steering of components identified via XAI offers a path toward actionable explanations, although its practical utility remains understudied. We introduce an interactive workflow combining SAE-based attribution with activation steering for instance-level analysis of concept usage in vision models, implemented as a web-based tool. Based on this workflow, we conduct semi-structured expert interviews (N=8) with debugging tasks on CLIP to investigate how practitioners reason about, trust, and apply activation steering. We find that steering enables a shift from inspection to intervention-based hypothesis testing (8/8 participants), with most grounding trust in observed model responses rather than explanation plausibility alone (6/8). Participants adopted systematic debugging strategies dominated by component suppression (7/8) and highlighted risks including ripple effects and limited generalization of instance-level corrections. Overall, activation steering renders interpretability more actionable while raising important considerations for safe and effective use.
%
%

\end{abstract}

\section{Introduction}

Unlike traditional engineered systems, deep neural networks develop their internal structures through optimization over billions of parameters rather than through deliberate architectural specification~\cite{amari_backpropagation_1993,goodfellow_qualitatively_2014,fazi_beyond_2021}. The semantic knowledge encoded by individual neurons, channels, or attention heads remains largely unknown~\cite{guidotti_survey_2018,das_opportunities_2020}, creating challenges in safety-critical applications where comprehending failure modes is essential~\cite{belle_principles_2021,nikiforidis_enhancing_2025}. The explainable AI (XAI) research community has produced diverse methodologies to explain machine learning (ML) models; spanning local attribution techniques~\cite{ribeiro2016should,lundberg2017unified,bach2015pixel}, counterfactual explanations~\cite{guidotti2024counterfactual}, and concept-based approaches~\cite{kim2018interpretability,achtibat2023attribution,bau2017network}.

These methods enable what we term \textit{correlational inspection}: practitioners can observe which components or features receive high attribution for a prediction, identifying candidates that may drive model behavior~\cite{holzinger2019causability}. However, precise identification of relevant components is only half the picture. To verify whether highly attributed components \textit{causally drive} a prediction rather than merely correlate with it, practitioners need interventive tools that act on the components explanations have surfaced~\cite{mansi2026evaluating,keenan2023mind}. Such tools remain largely unavailable in practice~\cite{mansi2026evaluating,cai2019human}.

Advances in mechanistic interpretability (MI) offer a path toward closing this gap, from correlational inspection to what we term \textit{actionable investigation}: directly manipulating model internals to test causal hypotheses about functional component roles. Sparse autoencoders (SAEs) can be applied to decompose model representations into interpretable components~\cite{dreyer2025mechanistic,dreyer2025attributing}, and attribution methods quantify each component's relevance to a given prediction. Activation steering complements this with an actionable capability by modifying these components to test their actual influence on model behavior~\cite{rimsky2024steering,panickssery2023steering,ghandeharioun2024s}. In principle, combining attribution with steering enables such actionable explainability previous work have called for~\cite{mansi2026evaluating,keenan2023mind}. However, existing work on steering has focused on dataset-level component selection~\cite{joseph2025steering}, system usability~\cite{li2025conceptviz}, and perceptibility of steering effects~\cite{diallo2025effectiveness} rather than its application for practitioners. Whether practitioners actually make the transition from correlational to causal reasoning when given steering access~\cite{bhalla2024towards}, and with what consequences for trust, strategy, and risk awareness, remains open.

\begin{figure*}
    \centering
    \includegraphics[width=0.9\linewidth]{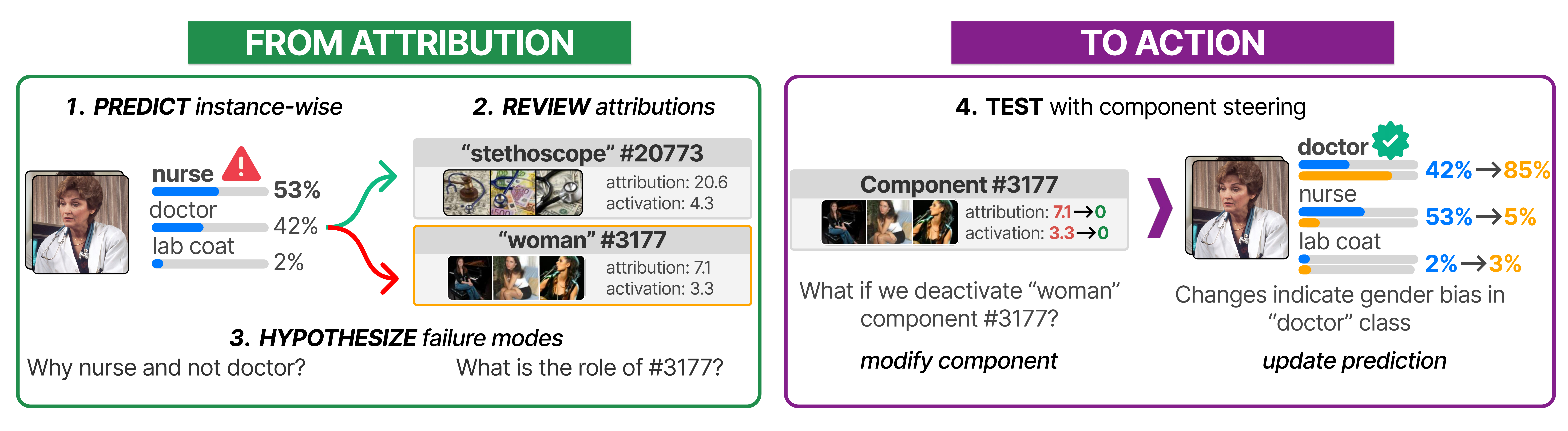}
    \caption{The four-step workflow from attribution to action: practitioners review component attributions, form causal hypotheses about components' roles, and test them via activation steering. The resulting prediction updates provide immediate feedback for hypothesis evaluation. Example shows suppressing the ``woman'' component to assess its influence on job role classification for ``nurse''.}
    \label{fig:placeholder}
\end{figure*}

In this work, we operationalize and evaluate the transition from attribution inspection to actionable investigation for instance-level model debugging. We structure this transition into a four-step workflow, from prediction review through attribution analysis and hypothesis formation to hypothesis testing via activation steering, and implement it in SemanticLens, a web-based tool for instance-level investigation of vision-language models. Using this workflow as experimental vehicle, we conducted semi-structured expert interviews with ML researchers and engineers (N~=~8) with two debugging tasks on CLIP~\cite{radford2021learning} to investigate whether the transition occurs in practice and what reasoning patterns, trust dynamics, investigation strategies, and perceived risks emerge.

\section{Related Work}
\paragraph{Mechanistic Foundation of Model Interventions}
The field of mechanistic interpretability seeks to reverse-engineer neural networks by decomposing them into understandable components and causal mechanisms~\cite{zhang2026locate,zhao2024explainability,geiger2025causal}. A recent survey~\cite{zhang2026locate} organizes mechanistic MI methods into a ``Locate, Steer, and Improve'' pipeline, categorizing localization techniques (including gradient-based attribution) and intervention approaches (including different steering methods). 

Model intervention can be applied for both understanding model representations and controlling model behavior. Early work on indiscriminate ablations~\cite{sejnowski1987parallel,smolensky1986neural} applied random interventions as a tool to understand general network properties, while later targeted ablations~\cite{meyes2019ablation,milliere2025interventionist} were used for testing functional hypotheses. More recently, activation and relevance patching~\cite{zhang2023towards,jafari2025relp}, also known as causal tracing~\cite{meng2022locating} and interchange intervention~\cite{geiger2021causal}, emerged as a method to isolate which components mediate specific computations by replacing activations from counterfactual inputs~\cite{milliere2025interventionist}.

Beyond model understanding, intervention has been applied to steer models towards desired behavior. Contrastive Activation Addition (CAA) enables steering of language model behavior via direction vectors~\cite{rimsky2024steering,panickssery2023steering,ghandeharioun2024s}.
Recent work demonstrates that SAE features in CLIP can be effectively steered to influence model outputs~\cite{joseph2025steering}, with approximately 10-15\% of features exhibiting meaningful steerability.

Bhalla et al.~\cite{bhalla2024towards} observe that while causal intervention has been employed to assess explanation faithfulness~\cite{mueller2024quest,belrose2023eliciting,chan2022causal,olah2020zoom}, these approaches rarely examine intervention as a method for practical control and debugging.

\paragraph{Interactive Visual Analytics for Model Validation and Error Analysis}
The goal of actionable model interventions has motivated development of several interactive visual analytics systems that combine explanation techniques with user-facing interfaces for systematic debugging. AttributionScanner~\cite{xuan2025attributionscanner} identifies data slices through attribution-based clustering without requiring metadata, while VISLIX~\cite{yan2025vislix} generates natural language explanations of error patterns using foundation models. SLIM~\cite{xuan2024slim} combines attention-weighted feature representations with human feedback to actively filter and rebalance datasets, and SUNY~\cite{xuan2024suny} analyzes necessity and sufficiency of learned features as causal explanations. These systems collectively establish that bridging correlational observation with actionable intervention through human feedback, data curation, or direct model steering is central to practical explainability.

\paragraph{Human-Centered Evaluation of Steering}
While the MI field has developed intervention techniques opening a new human-AI interaction space for model understanding and control, their human evaluation remains limited. To our knowledge, only two works~\cite{li2025conceptviz,diallo2025effectiveness} have measured user-facing outcomes.

ConceptViz~\cite{li2025conceptviz} is a visual analytics system for LLM concept exploration and validation with activation steering. The authors conduct a user study (N~=~12) to evaluate the effectiveness of ConceptViz system components in supporting users to explore and understand features as well as the overall system usability and workflow. The evaluation focuses on interface and high-level usability aspects without investigating the impact of direct model intervention with steering on users' mental models and trust.

Diallo et al.~\cite{diallo2025effectiveness} conduct the most comprehensive human evaluation of steering to date (N~=~190), measuring whether users perceive emotion control in language model outputs. In a user study, participants rate perceived emotional intensity and text comprehensibility across six emotions and eight steering strengths. Results demonstrate that steering successfully amplifies target emotions, with significant main effects of steering strength across five emotions. However, this evaluation measures perceptibility (i.e., can users detect the steering effect?) and identified a quality tradeoff (i.e., steering strengths beyond a threshold progressively degrade coherence), but does not assess whether emotion control improves task performance, trust calibration, or decision quality.

The related field of XAI offers established frameworks for human-centered evaluation that can inform steering assessment~\cite{doshi2017towards,lage2019human,kim2024human,naveed2024overview}. Doshi-Velez \& Kim~\cite{doshi2017towards} define functionally-grounded evaluation, where ``real humans perform real tasks'' to assess whether explanations improve decision quality. Building on this, Lage et al.~\cite{lage2019human} propose measuring trust calibration, cognitive workload, and behavioral patterns as core metrics for a human evaluation of interpretability. Empirical work demonstrates that explanations can both improve~\cite{bansal2021does} and harm~\cite{zhang2020effect} decision quality depending on their fidelity, underscoring the need to measure human outcomes.

\section{Investigating Model Predictions via Steering}
This section operationalizes the transition from attribution to action as an interactive workflow for instance-level model investigation. We first describe the technical foundation that enables component attribution and steering, building on prior work on SAE-based decomposition and attribution~\cite{dreyer2025attributing}, and then follow with the workflow.

\subsection{Technical Foundation}
Individual neurons in transformers are often polysemantic~\cite{elhage2022toy,milliere2025interventionist}. Following Dreyer et al.~\cite{dreyer2025attributing}, we train SAEs to extract monosemantic latent components from CLIP's representations. Each latent CLIP embedding $\mathbf{x}$ is decomposed as:
\[
\mathbf x = \sum_{j=1}^{d_{\text{SAE}}} a_j(x) \mathbf v_j + \mathbf b + \boldsymbol  \epsilon(\mathbf x),
\]
where $a_j(x)$ denotes the activation of component $j$ with feature direction $\mathbf v_j$, $b$ is a bias term, and $\epsilon(x)$ represents reconstruction error. To assign human-interpretable descriptions, we compute semantic alignment scores against textual labels $t \in \mathcal{T}$ using the average visual embedding $\bar{\boldsymbol x}_j$ of the top-$k$ most activating samples:
\[
s_j(\boldsymbol t) = \frac{\bar{\boldsymbol x}_j \cdot \boldsymbol t}{\|\bar{\boldsymbol x}_j\|\|\boldsymbol t\|} - \frac{\bar{\boldsymbol x}_j \cdot \boldsymbol t_{\text{empty}}}{\|\bar{\boldsymbol x}_j\|\| \boldsymbol t_{\text{empty}}\|},
\]
where $\boldsymbol t_{\text{empty}}$ is the embedding of an empty prompt. We quantify each component's influence on predictions through instance-wise attribution using Activation$\times$Gradient (resembling Input$\times$Gradient~\cite{shrikumar2017learning}):
\[
R_j(\boldsymbol x,\boldsymbol t) = a_j \frac{\partial y(\boldsymbol x,\boldsymbol t)}{\partial a_j},
\]
for a given image-text (embedding) pair $(\boldsymbol x, \boldsymbol t)$ and model output $y(\boldsymbol x,\boldsymbol t)$, providing two key insights per component: its semantic meaning via $s_j(\boldsymbol t)$ and its relevance to the current prediction via $R_j(\boldsymbol x,\boldsymbol t)$.

Based on this information, users can formulate and test causal hypotheses about component roles through interactive steering. For a selected SAE component $j$ with original activation $a_j$, a continuous control parameter $m_j \in [-1,1]$ rescales the activation as:
\[
a'_j = a_j(1+m_j),
\]
where $m_j = -1$ suppresses the component entirely \mbox{($a'_j = 0$)}, $m_j = 0$ leaves it unchanged, and $m_j = 1$ doubles its activation ($a'_j = 2a_j$). Unlike prior work that selects features based on dataset-level activation statistics~\cite{joseph2025steering}, our approach uses instance-wise attribution scores $R_j(x,t)$ to identify components most relevant to specific predictions. Users select components for steering based on their attribution scores and semantic descriptions, forming a set $\mathcal{S}=\{j_k\}_{k=1}^{n}$ with corresponding steering values $\{m_{j_k}\}_{k=1}^{n}$. Re-running inference with these modified activations reveals prediction changes.

Following the locate-and-steer paradigm described by Zhang et al.~\cite{zhang2026locate}, our approach combines activation-weighted gradient attribution for localization and amplitude manipulation for steering. The continuous parameter $m$ enables exploring dose-response relationships between component activation and prediction outcome, allowing users to test whether highly-attributed components causally drive instance-level predictions.

\subsection{From Attribution to Action: Workflow} \label{subsec:modelinteraction}

\begin{figure*}[!t]
 \centering
 \includegraphics[width=0.9\textwidth]{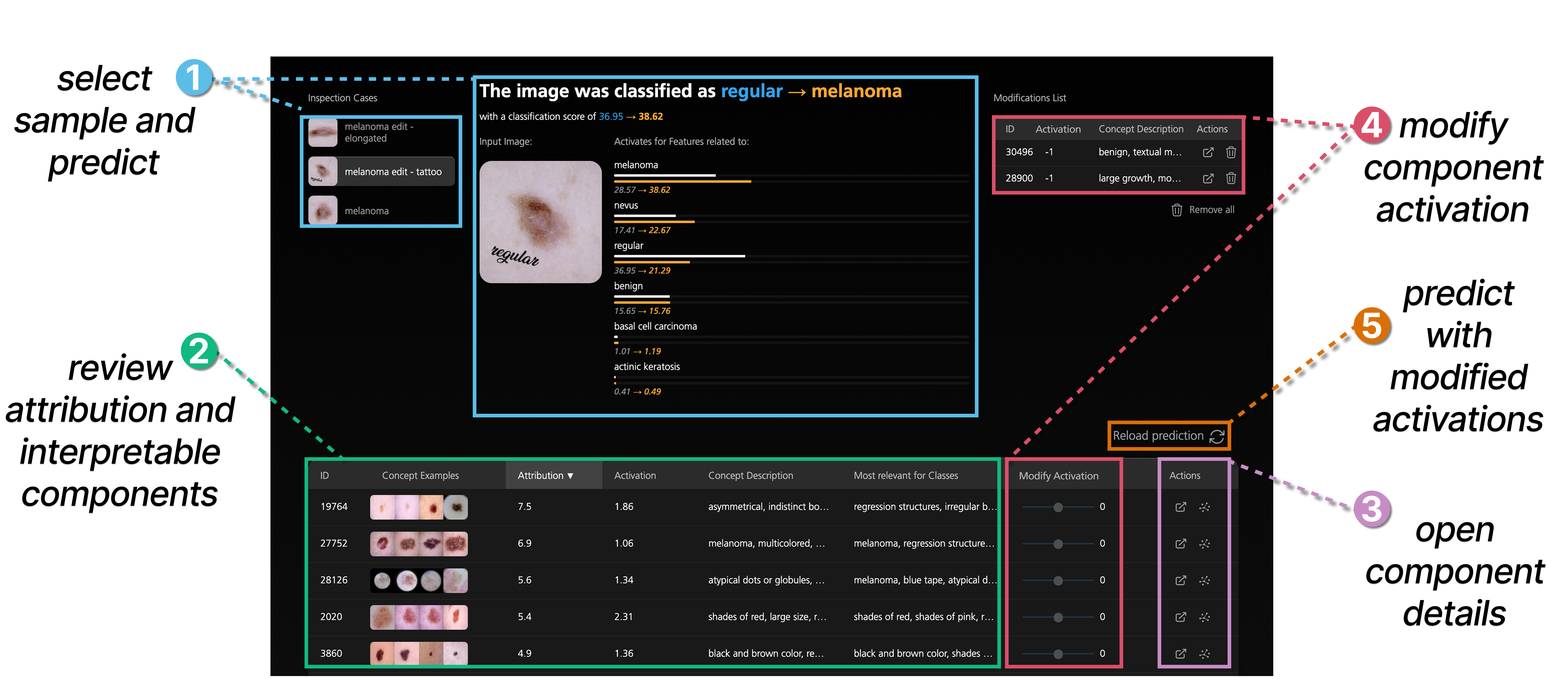}
  \caption{SemanticLens workflow implementation. Users select inspection samples (1), review components ranked by attribution (2), examine component details and visualizations (3), apply steering modifications (4), and observe prediction changes (5). Yellow text and bars indicate post-modification state. Example shown: a melanoma case corrected after suppressing component \#30496, which responds to text artifacts.}
 \label{fig:model_details}
\end{figure*}

We implement the technical components described above as an interactive workflow in SemanticLens, a web-based tool for instance-level model investigation (\autoref{fig:model_details}). The workflow guides users through four steps: (1)~prediction review, (2)~attribution analysis, (3)~hypothesis formation, and (4)~hypothesis testing through steering.

\textbf{Step 1: Predict.} The workflow begins with users selecting an input image for analysis. The system performs real-time inference, presenting the predicted class and confidence scores. Instance-wise attribution scores $R_j(x,t)$ are computed for all model components, establishing the foundation for exploration.

\textbf{Step 2: Review.} Components are ranked by their attribution and displayed in an interactive table (\autoref{fig:model_details}). For each component $j$, users can inspect: (i) its attribution $R_j(x,t)$ indicating influence on the current prediction, (ii) its activation value $a_j$, (iii) its semantic description derived from alignment scores $s_j(t)$, and (iv) highly activating example images revealing what visual patterns trigger the component. This step enables correlational inspection: practitioners can identify which components are associated with a prediction and what they encode, but cannot yet act on this information to test whether these associations are causal.

\textbf{Step 3: Hypothesize.} By examining attribution rankings alongside semantic descriptions, users can formulate hypotheses about model behavior. For instance, when a component shows high attribution for a melanoma classification and its visualization indicates ``textual markings'', a user might hypothesize that the model is vulnerable to typographic attacks. However, attribution alone cannot confirm this hypothesis, since highly attributed components may correlate with but not cause the prediction. Transitioning from attribution to action requires a final verification step.

\textbf{Step 4: Test.} Users test their hypotheses through targeted interventions by adjusting activations via $m_{j_k}$ for selected components $\mathcal{S}=\{j_k\}_{k=1}^{n}$. After defining modifications, users re-run the prediction and observe how outputs change, revealing whether the targeted components causally influence the prediction. This step closes the gap between inspection and action: correlational inspection of component attributions becomes actionable investigation of the causal roles of components.

\autoref{fig:model_details} illustrates this workflow for a typographic attack on the medical WhyLesionCLIP model~\cite{yang2024textbook,hufe2025dyslexify}: attribution analysis reveals a text-responsive component (\#30496) as the top contributor to a misclassification induced by overlaying the word ``regular''. Setting $m_{30496} = -1$ suppresses this component and reverts the prediction to the correct class. The implementation is publicly available as part of the SemanticLens webapp\footnote{\url{https://semanticlens.hhi-research-insights.eu}}.

The workflow structures the transition from attribution inspection to actionable investigation, but how practitioners apply it, and with what epistemic consequences, remains open.

\section{Qualitative Evaluation: Expert Interviews}\label{sec:evaluation}

To investigate how practitioners understand and apply activation steering, we conducted semi-structured expert interviews. Our evaluation addresses four research questions:

\begin{itemize}[leftmargin=*, nosep]
  \item\textbf{RQ1} How does the transition from attribution inspection to action affect practitioners' reasoning about model failures?
  \item\textbf{RQ2} How does actionable investigation affect the epistemic basis, the grounds on which practitioners justify their trust, and the perceived utility of explanations?
  \item\textbf{RQ3} What investigation strategies emerge when practitioners are given steering capabilities?
  \item\textbf{RQ4} What risks and limitations of the approach do practitioners perceive?
\end{itemize}

We investigate these questions using the workflow and implementation described in \autoref{subsec:modelinteraction} as the experimental setting. While the workflow provides the operational structure, our research questions and expert interviews target the epistemic effects of activation steering.

\subsection{Interview Design}
\begin{figure*}
    \centering
    \includegraphics[width=0.80\textwidth]{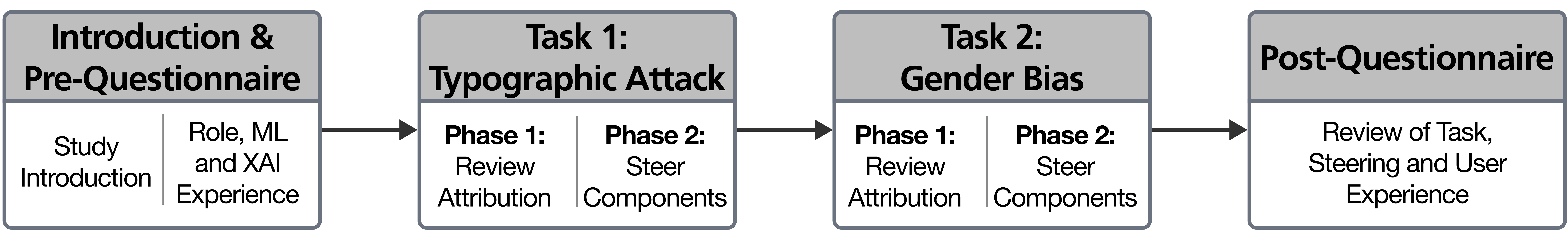}
    \caption{A process diagram of the interview structure: Pre-questionnaire, two debugging tasks each with attribution-only phase (Phase 1) and steering phase (Phase 2), and post-questionnaire.}
    \label{fig:study_process}
\end{figure*}

Each session lasted approximately 40~minutes and followed a think-aloud 
protocol~\cite{eccles2017think}. The procedure is illustrated in 
\autoref{fig:study_process}.

First, a pre-study questionnaire captured participants' professional roles, backgrounds, and prior experience with model debugging, bias detection, and explanation methods.

\begin{figure*}
    \centering
    \includegraphics[width=0.80\textwidth]{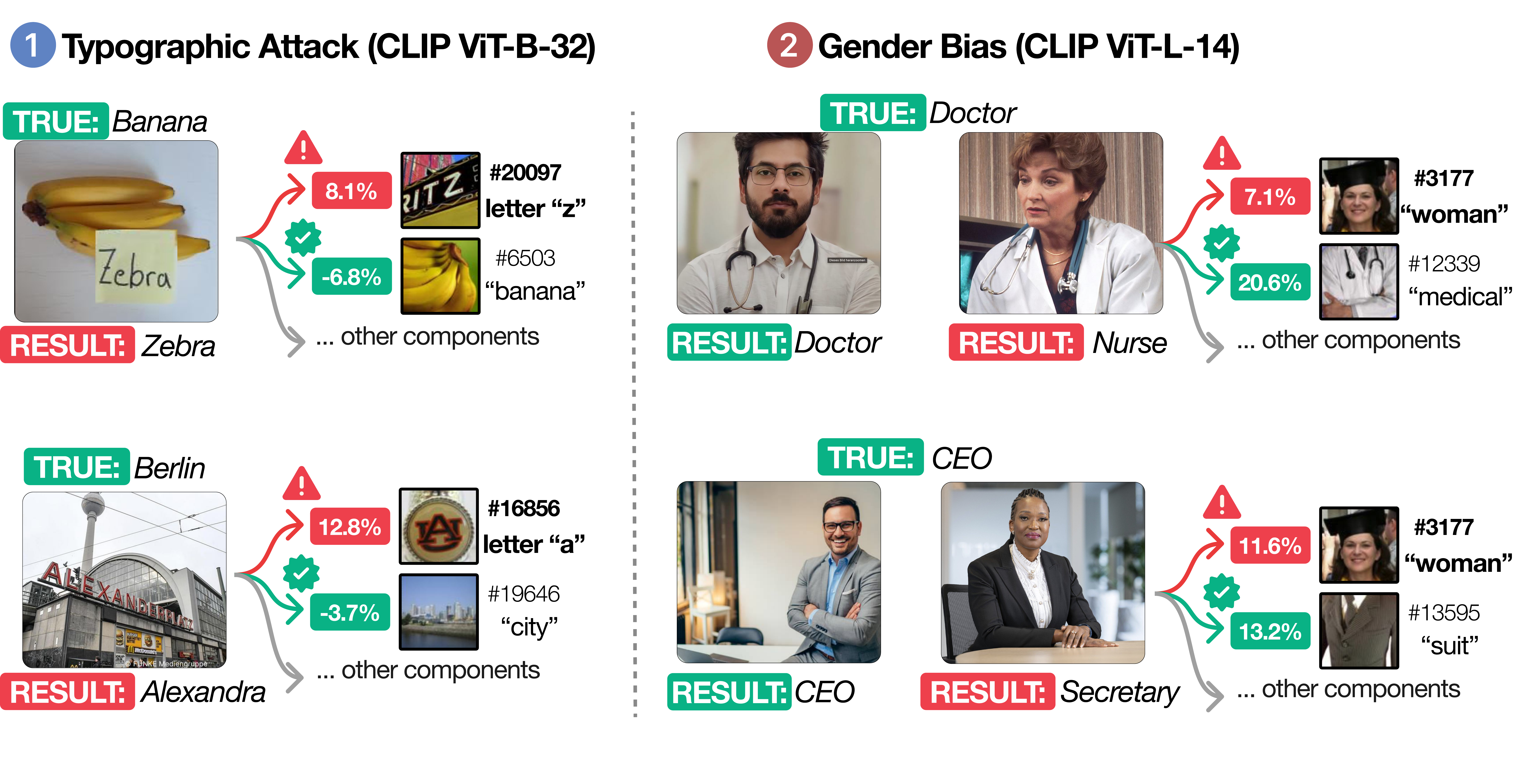}
    \caption{Two debugging tasks: Task 1 (typographic attack on CLIP ViT-B-32, where overlaid text causes misclassification~\cite{westerhoff2025scam}) and Task 2 (gender bias in CLIP ViT-L-14, where gender swap shifts predictions toward lower-status professions). Red badges indicate components contributing to misclassification; green badges indicate components supporting correct class. Component \#3177 (``woman'') appears among top-attributed features in both tasks.}
    \label{fig:study_tasks}
\end{figure*}

Second, participants completed two debugging tasks in SemanticLens (\autoref{fig:model_details}). Participants received a brief introduction explaining that it was an image classification task with the true label shown alongside each case, and were asked to explore the available inspection cases. Task~1 targeted a typographic vulnerability on CLIP ViT-B-32 from Westerhoff et al.~\cite{westerhoff2025scam}, where visible text overrides visual classification. Task~2 targeted a gender bias for job role classification in CLIP ViT-L-14, where swapping a male for a female person shifts predictions toward lower-status professions. Both tasks contained two inspection cases each (\autoref{fig:study_tasks}).

Each task consisted of two phases. In Phase~1, participants explored attributions and explanations without access to steering. This phase continued until participants either identified incorrect predictions and began forming hypotheses about root causes, or ceased to make further progress.  In Phase~2, steering controls were enabled, allowing participants to manipulate component activations and observe the effect on predictions. Semi-structured questions probed participants' understanding, confidence, trust, and strategy at the end of each phase. 

This sequential design was chosen to observe within-subject shifts in reasoning: by first exposing participants to attribution alone, we establish a baseline against which the effect of adding actionability in the form of steering can be assessed. As a consequence, observed shifts in reasoning coincide with, but cannot be causally attributed solely to, the introduction of steering. Other factors such as task familiarity, increased interface comfort, and general learning may contribute to the patterns we report.

Finally, a post-study questionnaire collected participants' reflections on the tasks, steering, and overall user experience.

\subsection{Study Participants}
We recruited eight ML researchers and engineers (P1--P8) with diverse expertise spanning vision-language models, diffusion models, weather/climate modeling, time-series analysis, and emotion recognition. Experience with XAI methods ranged from extensive (SAEs, feature-attribution methods, probing classifiers) to none. One participant (P1) had prior experience with activation steering; the remaining seven encountered it for the first time during the study.

\subsection{Method and Tools}

Sessions were recorded and transcribed via Microsoft Teams and anonymized prior to analysis. Responses were coded from transcriptions and supplementary handwritten interviewer notes. Interviews were conducted in English or German depending on participant preference and native language; German responses were translated by the first author whose native language is German. The study design was reviewed and approved by the Ethics Council of Fraunhofer Heinrich-Hertz-Institut. Given the exploratory nature, limited sample size, and novelty of human-centered evaluations of steering, our results focus on recurring themes and patterns rather than statistical analysis.

We employed thematic analysis with a deductive coding framework~\cite{clarke2017thematic}. A deductive codebook of 22~codes organized under the four research questions was developed from the interview protocol and iteratively refined during initial coding. Codes capture reasoning patterns (correlational vs.\ causal), trust bases (plausibility vs.\ evidence vs.\ conditional), debugging strategies (suppression, amplification, exploration), and perceived risks. Initial candidate codes were generated with LLM assistance (see \autoref{sec:genai}) from the interview transcripts and the pre-defined research questions. The authors then reviewed all candidate codes against the transcripts, merging overlapping codes, discarding codes that lacked grounding in the data, and relabeling codes for conceptual precision. This refinement reduced the initial set and consolidated it into the final 22~codes reported in \autoref{tab:coding_overview}. All codes were subsequently applied to the transcripts by the authors.

\begin{table}[t]
\centering
\caption{The 22 codes used to analyze interview data, organized by research question. \textit{Count} indicates the number of participants for whom each code was identified.}
\label{tab:coding_overview}
\small
\setlength{\tabcolsep}{4pt}
\begin{tabular}{@{}llc@{}}
\toprule
\textbf{RQ} & \textbf{Code} & \textbf{Count} \\
\midrule
\multirow{4}{*}{\rotatebox[origin=c]{0}{\parbox{0.7cm}{\centering \textbf{RQ1}}}}
 & Correlational reasoning (pre-steering) & 8 \\
 & Causal hypothesis formation & 6 \\
 & Causal test completed & 8 \\
 & Causal limitation acknowledged & 5 \\
\midrule
\multirow{3}{*}{\rotatebox[origin=c]{0}{\parbox{0.7cm}{\centering \textbf{RQ2}}}}
 & Plausibility-based trust & 2 \\
 & Evidence-based trust (post-steering) & 6 \\
 & Conditional trust & 3 \\
\midrule
\multirow{4}{*}{\rotatebox[origin=c]{0}{\parbox{0.7cm}{\centering \textbf{RQ3}}}}
 & Suppression strategy (necessity test) & 7 \\
 & Amplification strategy (sufficiency test) & 2 \\
 & Semantic scanning & 7 \\
 & Exploratory (no prior hypothesis) & 2 \\
\midrule
\multirow{4}{*}{\rotatebox[origin=c]{0}{\parbox{0.7cm}{\centering \textbf{RQ4}}}}
 & Ripple effects / non-orthogonality & 3 \\
 & Insufficient instance-level validation & 3 \\
 & Over-steering / performance degradation & 2 \\
 & Modification accumulation confounds & 2 \\
\bottomrule
\end{tabular}
\end{table}

\subsection{Task Outcomes}
\label{subsec:outcomes}

\autoref{tab:task_outcomes} summarizes behavioral outcomes across both tasks. Seven of eight participants identified the typographic attack and all eight identified the gender bias. Critically, all participants successfully corrected both failure modes through steering, including cases where participants had not identified the specific failure mode or responsible component beforehand.

\begin{table}[t]
\centering
\caption{Task outcomes across participants. \emph{Recog.}~= recognized the failure mode; \emph{Comp.}~= identified the responsible component; \emph{Fix}~= successfully corrected the prediction via steering; \emph{Att.}~= number of steering attempts to achieve fix. T1 = typographic attack task, T2 = gender bias task.}
\label{tab:task_outcomes}
\small
\setlength{\tabcolsep}{3pt}
\begin{tabular}{@{}l cc cc cc cc@{}}
\toprule
 & \multicolumn{4}{c}{\textbf{T1: Typographic}} & \multicolumn{4}{c}{\textbf{T2: Gender Bias}} \\
\cmidrule(lr){2-5} \cmidrule(lr){6-9}
\textbf{ID} & Recog. & Comp. & Fix & Att. & Recog. & Comp. & Fix & Att. \\
\midrule
P1 & \checkmark & \checkmark & \checkmark & 1 & \checkmark & \checkmark & \checkmark & 1 \\
P2 & \checkmark & \ding{56} & \checkmark & 2 & \checkmark & \checkmark & \checkmark & 1 \\
P3 & \checkmark & \checkmark & \checkmark & 2 & \checkmark & \checkmark & \checkmark & 1 \\
P4 & \checkmark & \checkmark & \checkmark & 3 & \checkmark & \checkmark & \checkmark & 1 \\
P5 & \checkmark & \checkmark & \checkmark & 1 & \checkmark & \checkmark & \checkmark & 3 \\
P6 & \checkmark & \checkmark & \checkmark & 3 & \checkmark & \checkmark & \checkmark & 1 \\
P7 & \ding{56} & \ding{56} & \checkmark & 4 & \checkmark & \checkmark & \checkmark & 2 \\
P8 & \checkmark & \checkmark & \checkmark & 1 & \checkmark & \ding{56} & \checkmark & 2 \\
\midrule
\textbf{Total} & 7/8 & 6/8 & 8/8 & \textit{M}~=~2.1 & 8/8 & 7/8 & 8/8 & \textit{M}~=~1.5 \\
\bottomrule
\end{tabular}
\end{table}

Two observations deserve emphasis. First, P7 corrected the typographic attack through four iterative steering attempts without having identified the problem or the responsible component, demonstrating that steering allowed the participant to explore model behavior simply by steering model components and responding to steering outcomes. Second, participants needed fewer attempts to correct the prediction via steering for T2 compared to T1 (\textit{M}~=~1.5 vs. \textit{M}~=~2.1). This likely reflects a learning effect, since T2 always followed T1, and may also be due to most participants perceived T2 as less difficult (see Table \ref{tab:task_outcomes}). 

The 100\% fix rate across both tasks constitutes a ceiling effect that limits the study's sensitivity to differences in tool effectiveness. Both failure modes were designed with known ground truth and involved perceptually salient cues (visible text overlay, gender swap), which eased identification and correction. These tasks served their intended purpose of enabling systematic observation of reasoning and strategy patterns during a short task, but the results should not be interpreted as evidence that the workflow would achieve similar success rates on complex tasks with subtler failure modes.

\subsection{RQ1: Shift in Reasoning Patterns}
\label{subsec:rq1}

During the attribution-only Phase~1, all participants engaged in correlational reasoning, trying to identify connections between components and predictions. For instance, P5 noted: \emph{``One can already guess that it is probably because the text was recognized''}.

After receiving steering access, all eight participants completed at least one causal test cycle: formulating or iteratively discovering a hypothesis, performing an intervention, and evaluating the outcome. Six participants (P1, P3--P6, P8) articulated explicit causal hypotheses before intervening. P4 exemplified this pattern for Task~2: \emph{``This concept should be less important, \ldots then it should jump to CEO''}; an assumption they confirmed by steering accordingly. P8 used the most explicitly causal wording of all participants. For Task~1, they stated: \emph{``Removing the text-related activation and then it works is a proof for the hypothesis''}. The remaining two participants (P2, P7) arrived at correct fixes through exploratory steering without articulating hypotheses beforehand, suggesting that steering can support discovery beyond hypothesis-driven investigation.

Notably, five participants (P2, P3, P5, P6, P8) acknowledged limitations of their causal claims without being prompted. P5 explicitly characterized the observed relationship as correlation rather than causation: \emph{``The interaction between the slider and the output behavior; a correlation was definitely there''}. P6 questioned whether components are truly independent: \emph{``\ldots whether the concepts are really orthogonal to each other or also influence other activations''}. This reflection suggests that steering enabled calibrated reasoning rather than inappropriately inflated causal confidence.

\subsection{RQ2: Trust and Utility Calibration}
\label{subsec:rq2}

We observed a change in the basis of trust reports after steering was introduced, though the sequential design precludes attributing this change solely to the steering capability. During the attribution-only phase, participants who expressed strong trust (2/8) grounded it in plausibility and coherence. P4 reported very high trust in the explanations, stating: \emph{``It was very consistent with what I saw and with the prediction''}. After steering, the same participant grounded trust in testability: \emph{``It is cool when you can directly change the influence and then check whether the model also changes the prediction, which is a kind of verification of my hypothesis''}.

Six of eight participants expressed evidence-based trust after steering, grounding their confidence in observed model responses to steering. Three participants additionally conditioned their trust on external validation. P2 stated that two tasks were insufficient and demanded full test-set metrics. P5 required robustness evaluation before considering safety-critical applications: \emph{``Medical applications where patient lives are at stake \ldots{} it would be too uncertain for me''}.

Regarding perceived utility of the explanation components, concept visualizations (example images showing what activates a component) were the most consistently valued element (7/8), while concept descriptions were frequently criticized as confusing (4/8). All participants assessed steering positively, emphasizing the speed and directness of feedback (P1, P4, P5), the live prediction update (P3, P6, P7), and the capacity to move beyond black-box perception (P8).

\subsection{RQ3: Investigation Strategies}
\label{subsec:rq3}

Before steering, participants used two complementary approaches to review components and build hypotheses. Most (7/8) used semantic scanning, inspecting concept visualizations and descriptions to identify suspicious components. Four of these participants additionally followed a top-down approach, starting from the highest-attributed component.

When steering, the dominant strategy was suppression of suspect components (7/8): participants tested whether removing a component's activation corrected the prediction. If suppressing a component restored correct classification, participants inferred that the component was responsible for the misclassification.

By contrast, only two participants tested through amplification of desired components. For example, P2 amplified banana- and city-related components to fix the predictions in the first task to ``Banana'' and ``Berlin''. P5 combined both approaches, first suppressing text features and then boosting banana features to fix the prediction to ``Banana'', while following the prediction outcomes and iteratively steering further: \emph{``Zebra is now only at 11\% down from 72\% \ldots{} now we are at banana at 80\%''}.

This asymmetry leads to design implications for SemanticLens: By ranking components by attribution to the (incorrect) prediction, the interface may implicitly guide users toward suppression strategies. 

\subsection{RQ4: Perceived Risks and Limitations}
\label{subsec:rq4}

Seven of eight participants reported at least one risk when asked whether they could imagine potential concerns. Notably, most participants (7/8) had no prior knowledge of steering and could therefore not base their assessment on technical specifics. The identified risks cluster into two categories.

\textbf{Technical risks} concern the mechanics of steering itself. Three participants (P2, P6, P8) raised the concern that modifying one component may have unintended effects on others if components are not independent. P8 proposed a concrete design solution: \emph{``If you make a change like this, if you could have some sort of global score \ldots{} being able to make a trade-off decision: is fixing this behavior locally worth the global effects?''}. Two participants (P1, P3) noted that excessive steering could potentially degrade overall model performance. These concerns align with established challenges in the MI literature around ripple effects~\cite{rinberg2025ripplebench}. 

\textbf{Methodological risks} concern the epistemological status of instance-level findings. Three participants (P1, P2, P5) emphasized that correcting individual predictions does not guarantee generalization. P1, the participant with prior steering experience, articulated the strongest standard: \emph{``The steering direction must generalize across different datasets''}. P5 explicitly cautioned against deployment in safety-critical domains without robustness guarantees. One participant (P3) questioned whether steering would provide value beyond the rather obvious cases from our study, requesting a task where they ``genuinely could not figure it out on their own''. This highlights a limitation of the present evaluation: both tasks involved failure modes with perceptually salient cues (visible text overlay, explicit gender swap), meaning that the root cause could often be hypothesized from visual inspection of the input alone. Whether the workflow and the observed reasoning patterns transfer to subtler failure modes, such as texture bias, frequency shortcuts, or spurious correlations not visible in the input, remains an open question that future evaluations should address with tasks where the failure mechanism is not apparent a priori.

\subsection{Additional Findings}
\label{subsec:additional}

Two further patterns emerged during the study with implications for future development of SemanticLens. Participants' mental models of the steering mechanism ($m \in [-1,1]$) diverged substantially: only one participant (P8) understood the multiplicative scaling mechanism (where $-1$ removes and $+1$ doubles activation), while others conceptualized steering as weight adjustment (P3, P4, P6, P7), fine-tuning (P2), or contrastive modification (P5). This heterogeneity suggests explicit interface communication through tooltips or mechanistic animations would reduce ambiguity. Furthermore, participants identified potential applications beyond CLIP debugging in climate attribution, model reparameterization, and time-series analysis, though some questioned whether discrete, interpretable concepts would emerge in non-vision domains.

\section{Conclusion}

We operationalize and evaluate the transition from inspection to action for instance-level model debugging through semi-structured expert interviews (N~=~8). All participants corrected both failure modes through steering, including cases resolved through exploratory steering without prior hypothesis formation. After steering was introduced, participants shifted from plausibility-based to evidence-based trust grounded in observed model responses, and most articulated explicit causal hypotheses while simultaneously acknowledging limitations of their causal claims. Strategy use was asymmetric, with suppression dominating over amplification, and participants surfaced concrete risks including ripple effects, over-steering, and insufficient instance-level validation. These findings provide first empirical evidence that activation steering, embedded in a structured workflow, can support the transition from correlational inspection to actionable investigation.

Several limitations warrant acknowledgment. The small expert sample (N~=~8) and perceptually salient failure modes limit generalizability to subtler defects also supported by the high success rates. The sequential design introduces potential learning confounds; counterbalancing was not pursued given the exploratory scope. The gap between instance-level correction and global model improvement requires dataset-level validation. Future work should extend evaluation to larger samples, subtler failure modes, and develop improved communication of the steering mechanism.
\newpage

\section*{Acknowledgements}
This work was supported by the Federal Ministry of Research, Technology and Space (BMFTR) as grants [BIFOLD (01IS18025A, 01IS180371I), xJuRAG (16IS25015B)]; the European Union’s Horizon Europe research and innovation programme (EU Horizon Europe) as grant ACHILLES (101189689); and the German Research Foundation (DFG) as research unit DeSBi [KI-FOR 5363] (459422098).

\section*{Declaration on the Use of Generative AI}\label{sec:genai}
During the preparation of this work, Claude Sonnet 4.5 and Grammarly were used for spelling and grammar checks, paraphrasing and rewording. Additionally, Claude Sonnet 4.5 was applied for generating initial qualitative codes during thematic analysis of the interviews. After using these tools, all authors reviewed and edited the content as needed and take full responsibility for the publication's content.

\section*{Ethical Statement}
The Ethics Commission of the Fraunhofer Heinrich-Hertz-Institut provided guidelines for the study procedure and approved the study design. Informed consent has been obtained from all participants.

{
    \small
    \bibliographystyle{ieeenat_fullname}
    \bibliography{main}
}


\end{document}